\documentclass[runningheads]{llncs}
\usepackage{graphicx}

\usepackage{tikz}
\usepackage{comment}
\usepackage{amsmath,amssymb} 
\usepackage{color}

\usepackage[accsupp]{axessibility}  

\usepackage{graphicx}
\usepackage{amsmath}
\usepackage{amssymb}
\usepackage{booktabs}
\usepackage{makecell}
\usepackage{warpcol}
\usepackage{enumitem}
\usepackage{algorithm}
\usepackage{algpseudocode}
\usepackage{multirow}
\usepackage{ulem}

\definecolor{citeCol}{RGB}{0,72,187}
\definecolor{linkCol}{RGB}{227,66,77}
\usepackage[pagebackref=false,breaklinks=true,colorlinks,urlcolor=magenta,citecolor=citeCol,linkcolor=linkCol,bookmarks=false]{hyperref}

\newcommand{\ke}[1]{{\color{black}#1}}

\begin{document}
\pagestyle{headings}
\mainmatter
\def\ECCVSubNumber{3203}  

\title{Harmonizer: Learning to Perform White-Box Image and Video Harmonization} 

\titlerunning{Harmonizer}
%
\author{
Zhanghan Ke\inst{1} \and
Chunyi Sun\inst{2} \and
Lei Zhu\inst{1} \and
Ke Xu\inst{1} \and 
Rynson W.H. Lau\inst{1}
}
\authorrunning{Z. Ke et al.}
%
\institute{Department of Computer Science, City University of Hong Kong \and
Australian National University
}
\maketitle

\begin{abstract}
Recent works on image harmonization solve the problem as a pixel-wise image translation task via large autoencoders. They have unsatisfactory performances and slow inference speeds when dealing with high-resolution images.
In this work, we observe that adjusting the input arguments of basic image filters, {\it e.g.}, brightness and contrast, is sufficient for humans to produce realistic images from the composite ones.
Hence, we frame image harmonization as an image-level regression problem to learn the arguments of the filters that humans use for the task. We present a \textit{Harmonizer} framework for image harmonization. 
Unlike prior methods that are based on black-box autoencoders, Harmonizer contains a neural network for filter argument prediction and several white-box filters (based on the predicted arguments) for image harmonization. We \ke{also} introduce a cascade regressor and a dynamic loss strategy for Harmonizer to learn filter arguments more \ke{stably} and precisely.
Since our network only outputs image-level arguments and the filters we used are efficient, Harmonizer is much lighter and faster than existing methods. Comprehensive experiments demonstrate that Harmonizer surpasses existing methods notably, especially with high-resolution inputs. Finally, we apply Harmonizer to video harmonization, which achieves consistent results across frames and 56 {\it fps} at 1080P resolution. 
Code and models are available at: {\color{blue}https://github.com/ZHKKKe/Harmonizer}.
\end{abstract}

 \begin{figure}[t]
\centering
\includegraphics[width=0.99\linewidth]{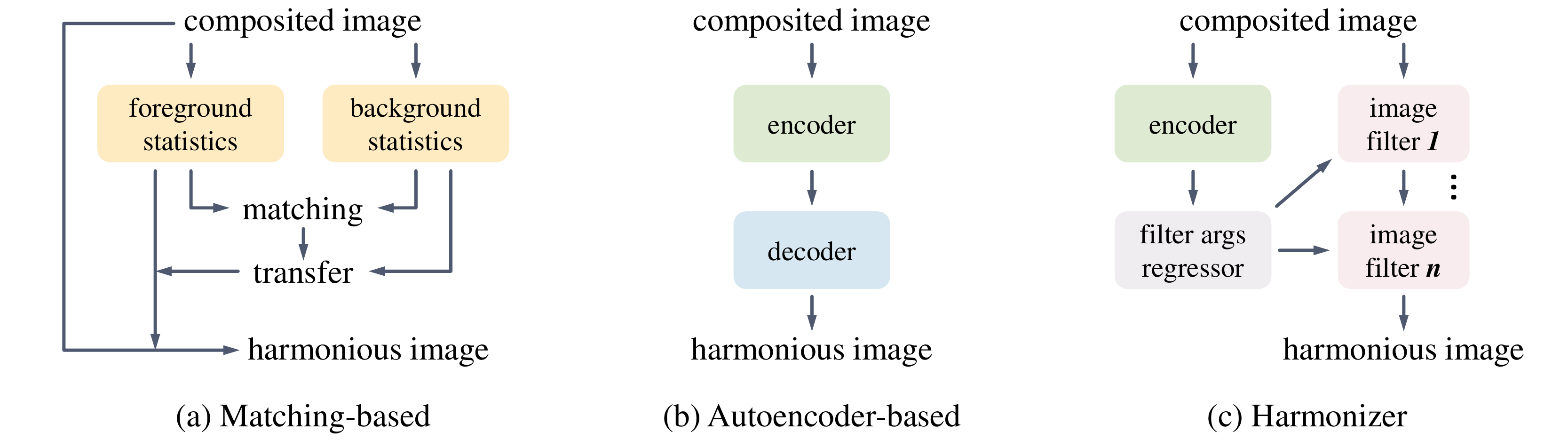}
{\begin{center}
\vspace{-0.5cm}
\caption{\textbf{Comparison of Harmonization Frameworks.} (a) Traditional matching-based methods transfer the background appearance to foreground regions based on hand-crafted statistics. (b) Autoencoder-based methods use black-box models to obtain harmonious images via pixel-wise image translation. (c) Our Harmonizer regresses image-level filter arguments to perform image harmonization in a white-box manner.}
\label{fig:framework}
\end{center}
}
\vspace{-0.9cm}
\end{figure}

\section{Introduction}
Extracting the foreground from one image and compositing it onto a background image is a popular operation in vision applications, \textit{e.g.}, image editing~\cite{IHGD2,PatchMatch} and stitching~\cite{ImageStitching1,ImageStitching2}. In order for the composite image to be more realistic, {\it i.e.}, cannot be easily distinguished by humans, the image harmonization task is introduced to remove the \ke{inconsistent} appearances between the foreground and background. This task is challenging because many conditions, such as lighting and imaging device being used, can affect object visual appearances~\cite{SFS,SIRS}, and humans are sensitive to even fine inharmony in appearances~\cite{IHHV1,IHCD4}. 

Traditional methods~\cite{Zhu_2015_ICCV,IHCD1,IHCT1,IHGD1,IHHV1,IHMS1,DPH} focus on matching the hand-crafted statistics between foreground and background regions, disregarding the semantic information which is vital for eliminating the large appearance gap. Recent deep-learning based methods~\cite{DIH,DoveNet,IHADFM,TVH,FASRIH,SSH,TransformerIH} leverage the strong semantic representation capability of autoencoders~\cite{UNet} to lower the appearance gap.
They regard image harmonization as a pixel-wise image translation task~\cite{pix2pix,pix2pixHD} from a \ke{composite} image to a harmonious version. Although they have achieved notable success, they also suffer from three key problems. 
First, their performances are unsatisfactory at high resolutions. Since using high-resolution images for training requires a huge amount of resources, these methods usually train and evaluate at low resolution.
Second, these methods are not suitable for mobile devices or real-time applications, due to their high computational overheads. The size of recent  autoencoder-based models~\cite{DoveNet,RAIN,BargainNet,IntrinsicIH} is larger than 100 MB, and their inference speed at 1080P (Full HD) resolution is only $\sim$10 {\it fps} on a RTX3090 GPU.  
Third, the images generated by these methods may not be consistent with the input images in terms of textures/details, {\it i.e.}, the original image contents may be changed, because neural networks are still essentially black-box
models. 

To design an efficient strategy for resolution-independent image harmonization in a white-box
manner, we conduct a user study to explore how humans perform this task. We observe that humans are able to produce realistic images by adjusting the input arguments of some basic image filters, such as brightness and contrast.
These filters also do not suffer from the three aforementioned problems, \textit{i.e.},
resolution-dependence, inefficiency, and black-box manner. Motivated by our observation, we formulate the image harmonization task as an image-level regression problem to learn the arguments of the filters used by humans, and present a \textit{Harmonizer} framework for the task. The key idea of our design is to combine a neural network and white-box filters for image harmonization, rather than just using black-box autoencoders. Specifically, in Harmonizer, the network contains a backbone encoder and a regressor for filter argument prediction, while the white-box filters use the predicted arguments to harmonize the input \ke{composite} images. Fig.\,\ref{fig:framework} summarizes the main differences between Harmonizer and existing frameworks.

\ke{To learn filter arguments more stably and precisely, we need to further consider two problems.}
First, the filter arguments are not easy to optimize simultaneously since they may affect each other. For example, if we adjust the brightness first before adjusting the highlight, we should consider the brightness argument when regressing the highlight argument. We note that utilizing a straightforward multiple-head regressor to predict each filter argument independently has unsatisfactory performances. 
\ke{We solve this problem by introducing a cascade regressor to predict each filter argument based on the features of the preceding filter arguments.}
Second, the loss of each filter output would accumulate all errors from the preceding filters, causing the regressor to bias towards some filters. We address this problem by introducing a dynamic loss strategy, which can balance the losses and helps Harmonizer pay more attention to the filters that are more difficult to learn.
\ke{Besides, we design a simple but effective exponential moving average (EMA) based strategy to adapt Harmonizer to video harmonization.}

We conduct extensive experiments to evaluate Harmonizer. The results on the iHarmony4 benchmark~\cite{DoveNet} demonstrate that Harmonizer outperforms prior state-of-the-art by a large margin. Harmonizer also has clear advantages in terms of model size and inference speed.
Our ablation study verifies the effectiveness of each component of Harmonizer. 
\ke{For video harmonization, Harmonizer obtains consistent results across frames
and an inference speed of 56 {\it fps} at 1080P resolution on a RTX3090 GPU.
}

\section{Related Works}

\subsection{Image Harmonization}

For an image composited of foreground image $F$ with foreground mask $M$ and background image $B$, the image harmonization task optimizes a harmonization function $\mathcal{H}$ that processes the foreground region $M F$ in order to match with the visual appearance of $B$, {\it i.e.}, creating a natural image $I$, as: 
\begin{equation}\label{eq:harmonization}
\begin{split}
I = \mathcal{H}( M F ) + (1 - M) B.
\end{split}
\end{equation}
Most traditional algorithms proposed $\mathcal{H}$ functions that concentrated on matching low-level appearance statistics, including color distributions~\cite{IHCD1,IHCD2,IHCD3,IHCD4}, color templates~\cite{IHCT1}, and gradient domain~\cite{IHGD1,IHGD2,IHGD3}. Some works further combined multi-scale statistics~\cite{IHMS1} or considered the visual realism of images~\cite{IHHV1,IHHV2}.

In recent years, many methods based on CNNs have been proposed with notable successes. These works regarded image harmonization as a pixel-wise image translation task, and their $\mathcal{H}$ functions are implemented based on autoencoders. For example, Tsai {\it et al.}~\cite{DIH} trained an end-to-end autoencoder to explore high-level semantics. Cun {\it et al.}~\cite{S2AM} introduced a spatial-separated attention module to leverage low-level appearances. Cong et al. [5,4] focused on finding more effective methods to guide the processing of the foreground using the information from the background.
Ling {\it et al.}~\cite{RAIN} related image harmonization with background-to-foreground style transfer. Guo {\it et al.}~\cite{IntrinsicIH} considered the intrinsic image characteristics to handle reflectance and illumination. Guo {\it et al.}~\cite{TransformerIH} replaced the CNN encoder with a Transformer to capture global background context. 

In spite of the success, all the aforementioned methods suffer from poor performances and slow inference speeds at high resolution, due to the low-resolution images used in the training process and the high computational overheads of autoencoders.
Instead, in this work, we formulate the image harmonization task as an image-level regression problem, and our proposed Harmonizer can solve the task with a consistent inference speed at high resolutions with negligible performance degradation.

\subsection{White-box Image Editing}

Recently, some works combined neural networks with human understandable ({\it i.e.}, white-box) algorithms for image editing. These methods usually have more stable performance than using only black-box neural networks. In addition, while the results from black-box neural networks may not be invertible, white-box algorithms allow users to further edit the images or undo any unwanted operations. For example, Yan {\it et al.}~\cite{Yan2016AutomaticPA} used model predictions to adjust pixel values.
Zou {\it et al.}~\cite{zou2020stylized} proposed a generative framework with a renderer/blender to simulate the human painting process. Hu {\it et al.}~\cite{Exposure} applied differentiable image operators for photo post-processing based on reinforcement learning. Wang {\it et al.}~\cite{LearnCartoonize} finished cartoon stylization by tuning the representations decomposed from the images.

In the image harmonization task, existing deep learning based methods are all based on black-box autoencoders~\cite{DIH,DoveNet,BargainNet,RAIN,IntrinsicIH,TransformerIH}, except for a concurrent work that attempts to support high-resolution inputs~\cite{concwork}. In contrast, our proposed Harmonizer combines a neural network with image filters to perform image harmonization in a white-box manner.

\ke{
The work most relevant to ours is probably Hu {\it et al.}~\cite{Exposure}. 
However, they used reinforcement learning to predict both types and arguments of filters. Besides, their method regresses only one filter in each step and may perform the same filter multiple times, resulting in a slow inference speed. 
Instead, we regress the arguments of a set of filters simultaneously and performs each filter only once, avoiding redundant filter operations and greatly improving efficiency.
}

\section{Harmonizer}

\subsection{Design Motivation}
Harmonizer aims to address image harmonization from a new perspective - combining neural networks with a white-box strategy. Since the white-box strategy that we select should be understood by humans, we first conduct a two-stage user study to analyze how humans perform image harmonization.

In the first stage, we investigate the white-box strategy humans use for image harmonization. We ask 5 experts who work in the image editing field (2 photographers, \ke{2} designer, and 1 painter) to process \ke{composite} images with Photoshop. We note that they accomplish this task \ke{mainly} by modifying some image properties through tools that can be split into a set of image filters. For example, the \ke{``Levels''} tool in Photoshop combines the highlight filter, the shadow filter, and the contrast filter. So, the 5 experts are essentially using image filters for image harmonization. In Harmonizer, we select the white-box strategy the same as the 5 experts: adjusting the arguments of appropriate filters to edit the foreground to match the background.

 \begin{figure}[t]
\centering
\includegraphics[width=0.99\linewidth]{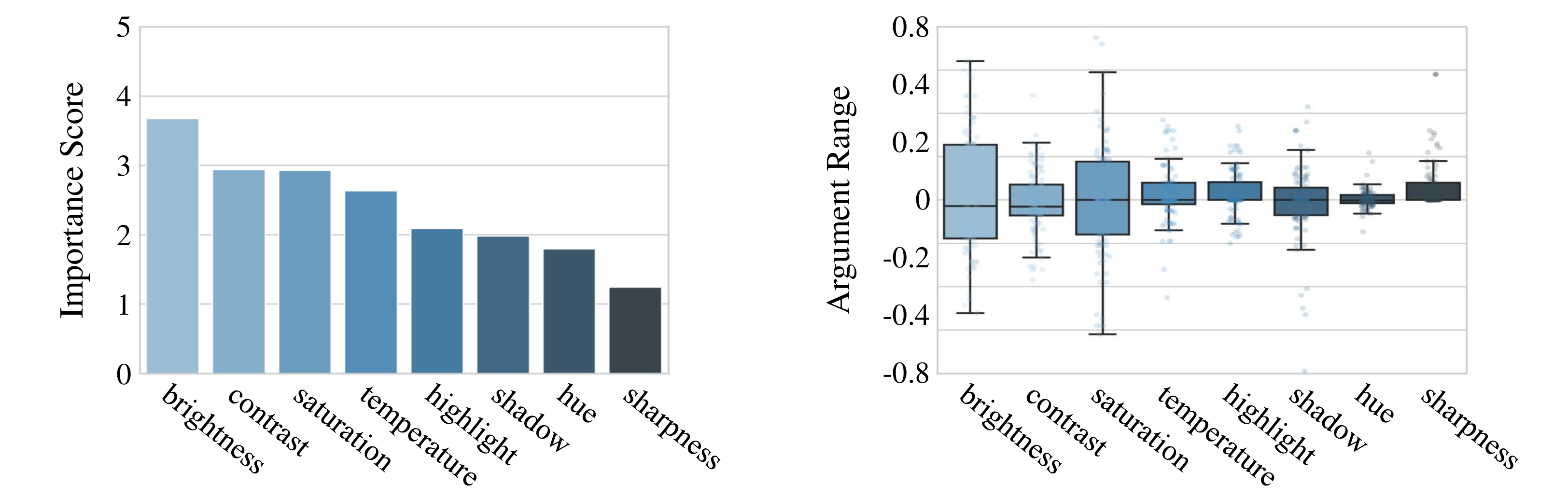}
{\begin{center}
\vspace{-0.5cm}
\caption{\textbf{Statistics of Our User Study.} {\it Left:} We sort image filters by their average importance scores. {\it Right:} For each filter, we use a Gaussian to fit its input arguments from the participants and visualize the argument distribution via Boxplots.}
\label{fig:userstudy}
\end{center}
}
\vspace{-0.7cm}
\end{figure}

In the second stage, we study the importance of different filters in humans' perception, and the value ranges that humans tune the filter arguments.
We build an image harmonization system based on the filters used in the first stage. We invite 26 participants. For each of them, our system will display 10 composite images, including 5 images that are identical among all participants and 5 images randomly selected for each participant. For each composite image, the participants are required to adjust the given filters to make it looks natural. Meanwhile, they should give an importance score for each filter, indicating its role in processing the \ke{composite} images. The score values are between 1 and 5. The higher the score, the more important the filter is. We record the importance scores and the filter arguments input by the participants for statistics. As shown in Fig.\,\ref{fig:userstudy}, the average importance scores of filters (Fig.\,\ref{fig:userstudy} {\it Left}) guide us to choose the filters with high scores, {\it i.e.}, the filters that are more important in humans' perception, for Harmonizer. The distributions of filter arguments (Fig.\,\ref{fig:userstudy} {\it Right})  guide us to set appropriate value ranges for filter arguments, {\it i.e.}, the value ranges that humans use.

Based on the user study above, we determine the white-box strategy used in Harmonizer (the first stage). We also understand which filters are important and the appropriate argument ranges for the filters (the second stage).

 \begin{figure}[t]
\centering
\includegraphics[width=0.99\linewidth]{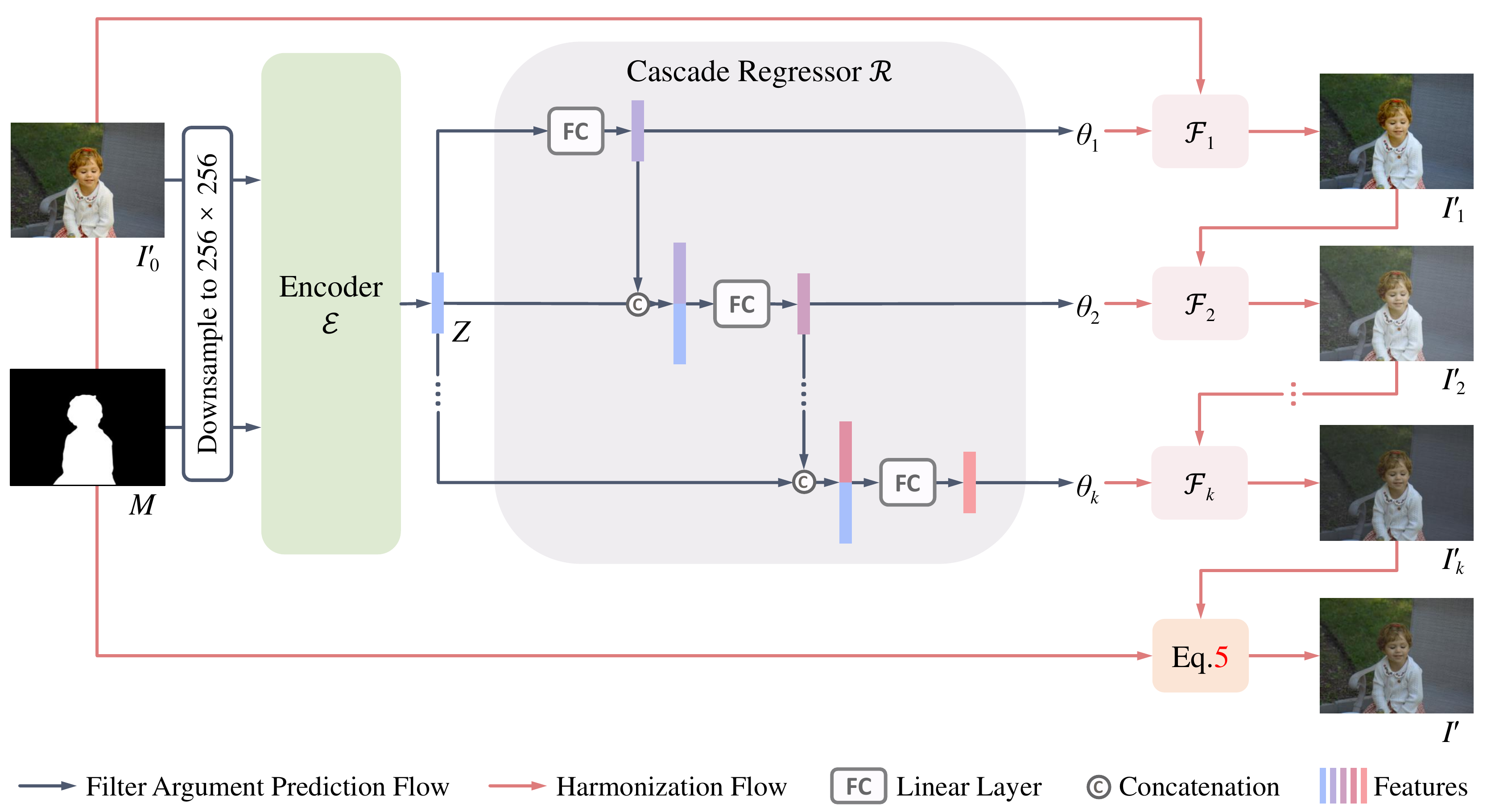}
{\begin{center}
\vspace{-0.5cm}
\caption{\textbf{The Harmonizer Framework.} For an input image $I^{\prime}_{0}$ with its foreground mask $M$, Harmonizer uses a neural network ($\mathcal{E}$+$\mathcal{R}$) to regress $k$ image filter arguments $\theta=\{\theta_{1}, \dots, \theta_{k}\}$ (\textit{i.e.}, the Filter Argument Prediction Flow). The image filters $\mathcal{F}=\{\mathcal{F}_{1}, \dots, \mathcal{F}_k\}$ in Harmonizer are then executed in sequence with the predicted arguments $\theta$ to obtain the 
output image $I^{\prime}$ (\textit{i.e.}, the Harmonization Flow).}
\label{fig:architecture}
\end{center}
}
\vspace{-0.7cm}
\end{figure}

\subsection{Architecture}\label{sec:architectural_details}
As shown in Fig.\,\ref{fig:architecture}, the framework of Harmonizer contains a backbone encoder $\mathcal{E}$, a regressor $\mathcal{R}$, and a set of image filters $\mathcal{F}=\{ \mathcal{F}_1, \dots, \mathcal{F}_k \}$, where $k$ indicates the number of filters.
The backbone $\mathcal{E}$ in Harmonizer is EfficientNet-B0\,\cite{EfficientNet}.
Given a composite image $I^{\prime}_{0}$ and its corresponding foreground mask $M$,
Harmonizer first downsamples them to the resolution of $256\times256$ and inputs them to $\mathcal{E}$ to extract image features $Z$ (with 160 channels), as:
\begin{equation}
    Z = \mathcal{E}(I^{\prime}_{0},\; M).
\end{equation}
Then, Harmonizer processes $Z$ by global pooling and uses $\mathcal{R}$ to regress filter arguments $\theta$ from it, as:
\begin{equation}
    \theta = \{\theta_1, \dots, \theta_k\} = \mathcal{R}(Z), \;\;\;\;\text{where}\;\;\;\; \theta_i \in [-1, 1],\;\; i=1, \dots, k.
\end{equation}
With $\theta$, Harmonizer executes the $k$ filters in sequence on $I^{\prime}_{0}$, as:
\begin{equation}
    I^{\prime}_{i} = \mathcal{F}_{i} (I^{\prime}_{i-1},\; \theta_{i}),\;\; i=1, \dots, k.
\end{equation}
Finally, the harmonious image $I^{\prime}$ is created by:
\begin{equation}\label{eq:harmonizer_final}
    I^{\prime} = M I^{\prime}_{k} + (1 - M) I^{\prime}_{0}.
\end{equation}
Eq.\,\ref{eq:harmonizer_final} ensures that the background regions in $I^{\prime}$ are the same as $I^{\prime}_{0}$, {\it i.e.}, the background pixels are not changed.

To balance the performance and the speed, we have also identified the preferred number of filters $k$ and which $k$ filters to use in Harmonizer. Our evaluations show that setting $k = 6$ is able to satisfy the real-time requirement (Table\,\ref{tab:ablation} {\it Right}). 
The six most important filters that we have selected based on human perception (Fig.\,\ref{fig:userstudy}\,\textit{Left}) for Harmonizer include brightness, contrast, saturation, color temperature, highlight, and shadow.

\textbf{Cascade Regressor.} 
Predicting the arguments for $k$ filters can be considered as a multi-task problem. One straightforward solution is to obtain each filter argument $\theta_i$ independently through a fully connected regressor $\mathcal{R}$ with $k$ heads $\{\mathcal{R}_{1}, \dots, \mathcal{R}_{k}\}$, as:
\begin{equation}\label{eq:multi-head-reg}
    \theta_{i} = \mathcal{R}_{i} (Z),\;\; i=1, \dots, k.
\end{equation}
\ke{However, Eq.\,\ref{eq:multi-head-reg} does not take into account the relationship between the filters.}
For example, both the brightness filter (with $\theta_{b}$) and the highlight filter (with $\theta_{h}$) will process the pixels with large pixel values. If we independently predict $\theta_{b}$ and $\theta_{h}$ from $Z$ \ke{and constrain them with ground truth, both of them will attempt to make the composite input image look harmonious}. As a result, the effects of these two filters will be accumulated in the output image, leading to an unsatisfactory result.  
To address this problem, we introduce a cascade regressor that uses the feature vector of the preceding filter arguments as conditions when regressing an argument $\theta_i$, as: 
\begin{equation}
\begin{split}
    &\theta_1 = \mathcal{R}_{1} (Z), \\
    &\theta_{i} = \mathcal{R}_{i} (Z \,|\, \theta_{i-1}) = \mathcal{R}_{i} (Z \,|\, \theta_{i-1}, \dots, \theta_{1}),\;\; i=2, \dots, k.
\end{split}
\end{equation}
In practice, after predicting a filter argument, we concatenate its feature vector with $Z$ to regress the next argument.

\subsection{Training Strategy}
We generate the composite input images from natural images for training.
Since Harmonizer executes the $k$ filters in a specific order $\mathcal{F}_{1}\to,\dots,\to\mathcal{F}_{k}$ on the composite image $I^{\prime}_0$, we reverse this filter order to $\mathcal{F}_{k}\to,\dots,\to\mathcal{F}_{1}$ to create $I^{\prime}_0$ from a natural image $I$, as:
\begin{equation}\label{eq:gt}
\begin{split}
    & I_{k} = I, \\
    & I_{k-i} = F_{k-i+1} (I_{k-i+1},\; \xi_{k-i+1}), \;\; i=1, \dots, k,\\
    & I^{\prime}_0 = I_0, \\
\end{split}
\end{equation}
where $\xi_{k-i+1}$ is the input arguments inside the range of $[-1, 1]$.
However, some filters ({\it e.g.}, the color temperature filter) are sensitive to the input arguments and may drastically change the image appearance with even a small change in argument value, resulting in an irreversible $I^{\prime}_0$, {\it i.e.}, we may not be able to recover $I$. 
To \ke{alleviate} this problem, we propose to reduce the range of the argument values
when generating $I^{\prime}_0$. As shown in Fig.\,\ref{fig:userstudy}\,\textit{Right}, our user study provides a rough argument range for each filter, which can guarantee the reversibility of the composite images \ke{in most cases}. Therefore, we sample the input argument $\xi_{i}$ for the filter $\mathcal{F}_{i}$ from a Gaussian distribution $\mathcal{G}_{i}$, as:
\begin{equation}
    \xi_{i} = \mathcal{G}_{i} (m_i, \, v_i), \;\; i=1, \dots, k,
\end{equation}
where $m_i$ and $v_i$ are the mean and variance from Fig.\,\ref{fig:userstudy}\,\textit{Right}, respectively.

\begin{figure}[t]
\centering
\includegraphics[width=0.99\linewidth]{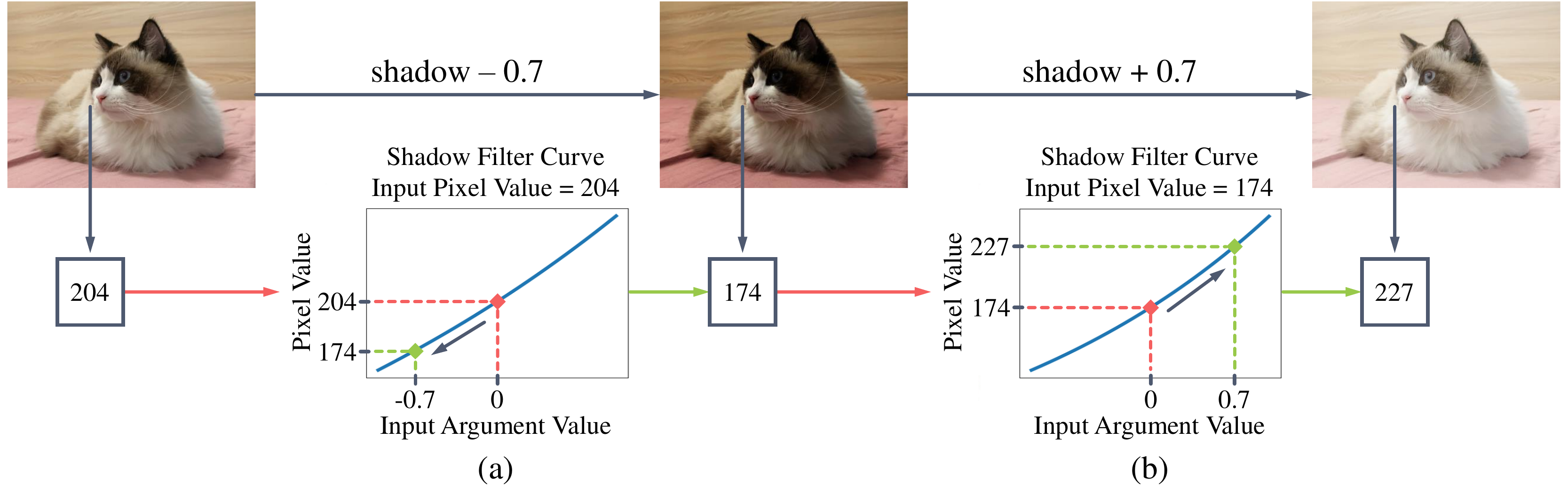}
{\begin{center}
\vspace{-0.5cm}
\caption{\textbf{Asymmetric Filter Operations.} For the shadow filter that we use, if we (a) adjust the shadow with an argument of -0.7, the pixel value of 204 will drop to 174. After that, if we (b) adjust the shadow with an argument of 0.7, the pixel value of 174 will increase to 227, which is not equal to the original pixel value of 204.}
\label{fig:shadow}
\end{center}
}
\vspace{-0.3cm}
\end{figure}

 \begin{figure}[t]
\centering
\includegraphics[width=0.99\linewidth]{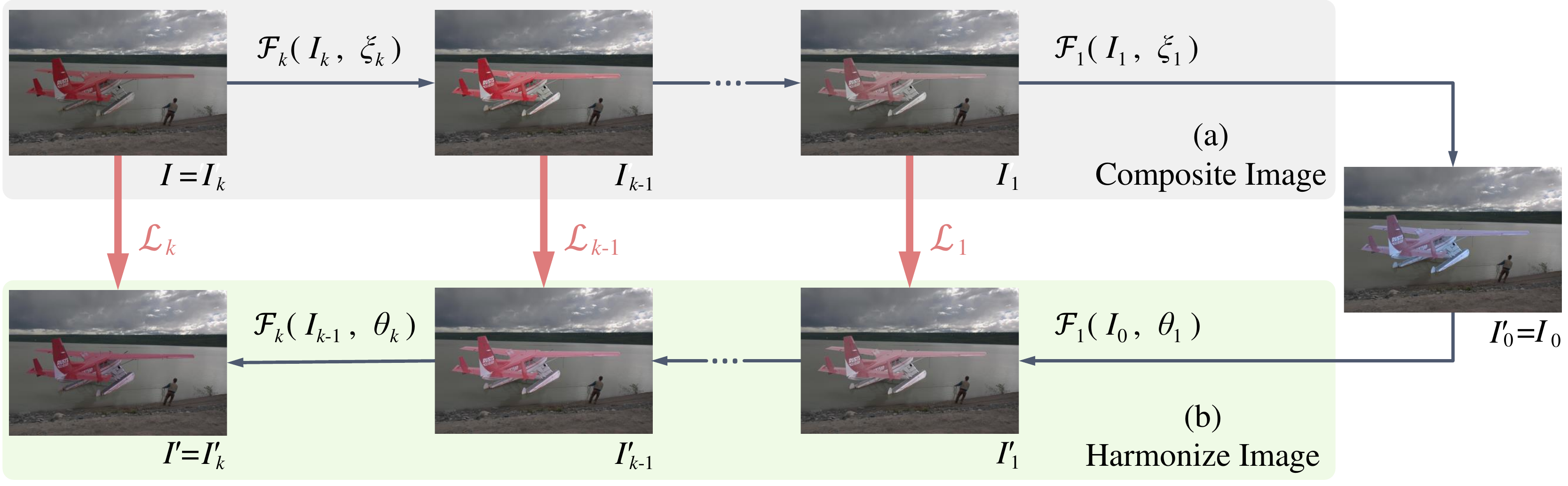}
{\begin{center}
\vspace{-0.5cm}
\caption{\textbf{Optimizing Filter Arguments $\theta$.} For (a) a composite image $I^{\prime}_{0}$ generated from a natural image $I$ using the filter arguments $\xi=\{\xi_1, \dots, \xi_k\}$, Harmonizer tends to (b) predict a set of arguments $\theta$ to recover a harmonious image $I^{\prime}$ from $I^{\prime}_{0}$. For some filters, the ground truth of $\theta_i$ is unknown. Hence, we optimize $\theta$ through the loss $\mathcal{L}_i$ between each filter output $I^{\prime}_{i}$ and its corresponding composite image $I_i$.}
\label{fig:loss}
\end{center}
}
\vspace{-0.7cm}
\end{figure}

Note that some of the filters used are non-linear, and their operations are asymmetric. Here, we take the shadow filter $\mathcal{F}_{s}$ as an example. As illustrated in Fig.\,\ref{fig:shadow}, if we use the shadow filter $\mathcal{F}_{s}$ with an input argument $\xi_s$ to adjust an image $I$, we may not recover $I$ using $\mathcal{F}_{s}$ with argument $-\xi_s$. Therefore:
\begin{equation}
    I \neq \mathcal{F}_s(\,\mathcal{F}_s ( \, I, \, \xi_s),\, -\xi_s).
\end{equation}
Hence, $-\xi_s$ cannot be used as the ground truth of the shadow filter argument $\theta_s$ predicted by Harmonizer. As shown in Fig.\,\ref{fig:loss}, instead of regressing $\theta$ directly, we optimize each $\theta_i$ through the L2 loss between the filter output $I^{\prime}_i$ and its corresponding composite image $I_i$ (calculated in Eq.\,\ref{eq:gt}), as:
\begin{equation}
    \mathcal{L}_i = M\; \|\,I^{\prime}_i\, - \,I_i\,\|_{2} = M\; \|\,\mathcal{F}_{i} (I^{\prime}_{i-1},\, \theta_i)\, - \,I_i\,\|_{2}, \;\; i=1, \dots, k.
\end{equation}
Here the foreground mask $M$ constrains the loss only on the foreground regions. 
We apply a loss on each output to ensure gradient propagation for the filters in the front. 
For the composite input image generated by GAN~\cite{GAN} (following \cite{DoveNet,BargainNet,TransformerIH} {\it etc.}) rather than Eq.\,\ref{eq:gt}, we only apply the loss $\mathcal{L}_{k}$ on the final output.

\textbf{Dynamic Loss Strategy.} During training, the Loss $\mathcal{L}_i$ usually increases with the filter index $i$ due to the inaccurate $I_{i-1}$ from the preceding filters, which may bias the regressor towards later filters. We introduce a dynamic strategy to balance the losses. We first subtract the errors accumulated by the preceding filters from $\mathcal{L}_i$. We then normalize $\mathcal{L}_i$ to enhance the loss of filters that introduce larger errors. Formally, the loss $\mathcal{L}_i$ is dynamically reweighted by:
\begin{equation}\label{eq:dynamic_loss}
    \mathcal{\tilde{L}}_i = max\Big(\frac{\mathcal{L}_i - \mathcal{L}_{i-1}}{\mathcal{L}_k}, \, 0\Big), \;\; i=1, \dots, k,
\end{equation}
\ke{Note that we detach the gradients at the denominator $\mathcal{L}_k$. If $\mathcal{L}_{i}<\mathcal{L}_{i-1}$, we reweight $\mathcal{L}_i$ to $0$ to focus on optimizing $\mathcal{L}_{i-1}$ as we consider $\mathcal{F}_{i}$ work well for the current input.}
The final training loss for Harmonizer is:
\begin{equation}\label{eq:total_loss}
    \mathcal{L} = \mu\,\sum^{k}_{i=1}\,\mathcal{\tilde{L}}_{i}.
\end{equation}
\ke{where $\mu$ is used to rescale $\mathcal{L}_i$ to ensure sufficient gradients for backpropagation.}

\subsection{Video Harmonizer}\label{sec:video}

Applying existing image harmonization algorithms individually on each video frame often leads to flickering of the foreground in the output sequence. Although some video processing methods~\cite{bonneel2013examplebased,bonneel2015blind,Lai-ECCV-2018,lei2020dvp} have been proposed to encourage the prediction consistency across video frames, they require a long processing time or additional modules for training. 
Therefore, obtaining stable results in real-time video harmonization is an unexplored problem.

We introduce here a simple but effective strategy for adapting Harmonizer to video harmonization. The idea behind our strategy is to ensure that the predicted filter arguments change smoothly across frames. 
We achieve this by smoothing the predicted arguments $\theta$ with exponential moving average (EMA), as:
\begin{equation}\label{eq:video}
    \bar{\theta}^{t} = (1 - \alpha) \, \bar{\theta}^{t-1} +  \alpha \, \theta^{t},
\end{equation}
where $t$ is the frame index, and $\alpha=0.9$ is an EMA coefficient.

\section{Experiments}

In this section, we first introduce the datasets, metrics, and training details for our experiments. 
We then compare Harmonizer with existing image harmonization methods (Sec.\,\ref{sec:4_1}).
We also show the effectiveness of adapting Harmonizer to video harmonization (Sec.\,\ref{sec:4_2}). 
We further conduct ablation experiments to evaluate the effectiveness of individual components in Harmonizer (Sec.\,\ref{sec:4_3}). 
Finally, we demonstrate the advantages of Harmonizer in real-world image/video harmonization applications through user studies (Sec.\,\ref{sec:4_4}).

\textbf{Datasets.} Following the recent works, we conduct our experiments on the iHarmony4 benchmark~\cite{DoveNet}, which contains four subsets: HCOCO, HAdobe5k, HFlickr, and Hday2night. Each sample in iHarmony4 consists of a natural image, a foreground mask, and a composite image (with the foreground generated by GAN\,\cite{GAN}). During training, we also create the composite images via Eq.\,\ref{eq:gt}. Note that this is a data augmentation method, without using any extra data. 

\textbf{Metrics.} We evaluate the image harmonization performance by Mean Square Error (MSE), foreground MSE (fMSE), and Peak Signal-to-Noise Ratio (PSNR). 
fMSE calculates MSE only on the foreground regions rather than the whole image, as image harmonization does not change the background appearance. 

\textbf{Training.} We train Harmonizer by Adam for 60 epochs. With a batch size of 16, the initial learning rate is set to $3e^{-4}$ and is multiplied by 0.1 after every 25 epochs. We set $\mu$ (in Eq.\,\ref{eq:dynamic_loss}) to 10. In all experiments, except the ablation on the number of filters, we use Harmonizer with the 6 filters stated in Sec.\,\ref{sec:architectural_details}.

\begin{table*}[t]
  \begin{center}
    \caption{\textbf{Quantitative Comparison on iHarmony4 at $256\times256$ Resolution.} All metrics are computed following the previous works. $\uparrow$ indicates the higher the better, while $\downarrow$ indicates the lower the better.}\label{tab:lowresolution}
\setlength{\tabcolsep}{3pt}
\scriptsize
\begin{tabular}{c|r|rrrrrr|r}
\toprule
Dataset & Metric & DIH\cite{DIH} & S2AM\cite{S2AM} & DOVE\cite{DoveNet} & BARG\cite{BargainNet} & IntrIH\cite{IntrinsicIH} & IHT\cite{TransformerIH} & \;\;\;\;\;\;\;\;Our \\
\midrule
\multirow{3}{*}{HAdobe5k}
& MSE$\downarrow$  &  92.65 &  63.40 &  52.32 &  39.94 &  43.02 &  47.96 &  \textbf{21.89} \\
& fMSE$\downarrow$ & 593.03 & 404.62 & 380.39 & 359.49 & 284.21 & 321.14 & \textbf{170.05} \\
& PSNR$\uparrow$   &  32.28 &  33.77 &  34.34 &  35.34 &  35.20 &  36.10 &  \textbf{37.64} \\
\midrule
\multirow{3}{*}{HFlickr}
& MSE$\downarrow$  & 163.38 & 143.45 & 145.21 &  97.32 & 105.13 &  88.41 &  \textbf{64.81} \\
& fMSE$\downarrow$ &1099.13 & 785.65 & 985.79 & 769.02 & 716.60 & 617.26 & \textbf{434.06} \\
& PSNR$\uparrow$   &  29.55 &  30.03 &  29.75 &  31.34 &  31.34 &  32.37 &  \textbf{33.63} \\
\midrule
\multirow{3}{*}{HCOCO}
& MSE$\downarrow$  &  51.85 &  41.07 &  36.72 &  24.84 &  24.92 &  20.99 &  \textbf{17.34} \\
& fMSE$\downarrow$ & 798.99 & 542.06 & 551.01 & 489.94 & 416.38 & 377.11 & \textbf{298.42} \\
& PSNR$\uparrow$   &  34.69 &  35.47 &  35.83 &  37.03 &  37.16 &  37.87 &  \textbf{38.77} \\
\midrule
\multirow{3}{*}{Hday2night}
& MSE$\downarrow$  &  82.34 &  76.61 &  56.92 &  50.98 &  55.53 &  58.14 &  \textbf{33.14} \\
& fMSE$\downarrow$ &1129.40 & 989.07 &1067.19 & 853.61 & 797.04 & 823.68 & \textbf{542.07} \\
& PSNR$\uparrow$   &  34.62 &  34.50 &  35.53 &  35.88 &  35.96 &  36.38 &  \textbf{37.56} \\
\midrule
\multirow{3}{*}{All}
& MSE$\downarrow$  &  76.77 &  59.67 &  52.36 &  37.82 &  38.71 &  37.07 &  \textbf{24.26} \\
& fMSE$\downarrow$ & 778.41 & 537.23 & 541.53 & 513.16 & 400.29 & 395.66 & \textbf{280.51} \\
& PSNR$\uparrow$   &  33.41 &  34.35 &  34.75 &  35.88 &  35.90 &  36.71 &  \textbf{37.84} \\
\bottomrule
\end{tabular}
\vspace{-0.5cm}
\end{center}
\end{table*}

\begin{table*}[ht]
  \begin{center}
    \caption{\textbf{Quantitative Comparison on iHarmony4 at High Resolutions.} All metrics are calculated at the original image resolution of the samples in iHarmony4. 
    The inputs to the existing methods are in low-resolution. Their outputs are then bilinearly upsampled to high resolutions for metric calculation. 
    We also apply Polynomial Color Mapping for upsampling (with subscript ``+PCM'').
    }\label{tab:highresolution}
\setlength{\tabcolsep}{3pt}
\scriptsize
\begin{tabular}{c|r|rrrrrr|r}
\toprule
Dataset & Metric & DOVE\cite{DoveNet} & DOVE\cite{DoveNet} & BARG\cite{BargainNet} & BARG\cite{BargainNet} & IHT\cite{TransformerIH} & IHT\cite{TransformerIH} & \;\;\;\;\;\;\;\;Our \\
& & & +PCM & & +PCM & & +PCM & \\
\midrule
\multirow{3}{*}{HAdobe5k}
& MSE$\downarrow$  &  68.16 &  72.08 &  77.96 &  88.20 &  56.90 &  63.28 &  \textbf{24.37} \\
& fMSE$\downarrow$ & 511.02 & 579.21 & 560.49 & 689.58 & 465.72 & 547.61 & \textbf{196.12} \\
& PSNR$\uparrow$   &  33.30 &  32.82 &  32.65 &  32.17 &  33.63 &  33.04 &  \textbf{37.80} \\
\midrule
\multirow{3}{*}{HFlickr}
& MSE$\downarrow$  & 172.80 & 159.46 & 159.34 & 150.67 & 135.49 & 127.10 &  \textbf{69.19} \\
& fMSE$\downarrow$ &1192.88 &1110.22 &1114.29 &1096.91 & 994.23 & 976.08 & \textbf{479.26} \\
& PSNR$\uparrow$   &  28.81 &  29.71 &  29.01 &  29.88 &  29.59 &  30.44 &  \textbf{33.37} \\
\midrule
\multirow{3}{*}{HCOCO}
& MSE$\downarrow$  &  56.49 &  47.13 &  52.84 &  46.62 &  44.95 &  40.16 &  \textbf{20.93} \\
& fMSE$\downarrow$ &1000.14 & 844.84 & 940.79 & 844.21 & 838.86 & 785.03 & \textbf{374.96} \\
& PSNR$\uparrow$   &  33.35 &  34.5 &  33.54 &   34.51 &  34.19 &  34.85 &  \textbf{37.69} \\
\midrule
\multirow{3}{*}{Hday2night}
& MSE$\downarrow$  &  58.23 &  67.81 &  53.99 &  66.91 &  63.26 &  72.94 &  \textbf{37.28} \\
& fMSE$\downarrow$ &1125.46 &1007.99 & 958.02 &1109.15 & 988.56 &1054.97 & \textbf{640.74} \\
& PSNR$\uparrow$   &  35.44 &  35.17 &  35.65 &  35.07 &  35.71 &  35.06 &  \textbf{37.15} \\
\midrule
\multirow{3}{*}{All}
& MSE$\downarrow$  &  72.98 &  67.35 &  71.93 &  70.76 &  58.89 &  57.22 &  \textbf{27.62} \\
& fMSE$\downarrow$ & 882.26 & 800.48 & 851.83 & 832.62 & 750.06 & 741.98 & \textbf{339.23} \\
& PSNR$\uparrow$   &  32.86 &  33.49 &  32.82 &  33.32 &  33.54 &  33.83 &  \textbf{37.23} \\
\bottomrule
\end{tabular}
\vspace{-0.2cm}
\end{center}
\end{table*}

\begin{table*}[t]
  \begin{center}
    \caption{\textbf{Comparison on Inference Speed, Model Size, and GPU Memory.} The speed evaluation is conducted at 1080P resolution on a RTX3090 GPU.}\label{tab:other}
\setlength{\tabcolsep}{5pt}
\scriptsize
\begin{tabular}{r|rrrrr|r}
\toprule
Metric & S2AM\cite{S2AM} & DOVE\cite{DoveNet} & BARG\cite{BargainNet} & IntrIH\cite{IntrinsicIH} & IHT\cite{TransformerIH} &\;\;\;\; Our \\
\midrule
Inference Speed ({\it fps}) $\uparrow$     & 6.76  & 13.8 & 11.6 & 1.2 & 5.1 & \textbf{56.3} \\
Model Size (MB) $\downarrow$              &  268.1  & 219.1 & 234.9 & 163.5 & 25.8 & \textbf{21.7} \\
GPU Memory (GB) $\downarrow$             & 6.3 & 6.5 & 3.7 & 16.47 & 18.5 & \textbf{2.3} \\
\bottomrule
\end{tabular}
\vspace{-0.5cm}
\end{center}
\end{table*}

\subsection{Comparison with State-of-the-arts}\label{sec:4_1}
We compare Harmonizer with recently proposed methods, including DIH~\cite{DIH}, S2AM~\cite{S2AM}, DOVE~\cite{DoveNet}, BARG~\cite{BargainNet}, IntrIH~\cite{IntrinsicIH}, and IHT~\cite{TransformerIH}. We use the pre-trained models released by their authors for evaluation. We first follow the prior works to evaluate all methods at low resolution, {\it i.e.}, the output harmonious images and the ground truths will be resized to $256\times256$ for metric calculation. As shown in Table\,\ref{tab:lowresolution}, Harmonizer outperforms the existing methods on all four subsets of iHarmony4. Notably, compared to the state-of-the-art method, Harmonizer reduces the average MSE across all subsets by $35\%$.

\begin{figure}[t]
\centering
\includegraphics[width=0.99\linewidth]{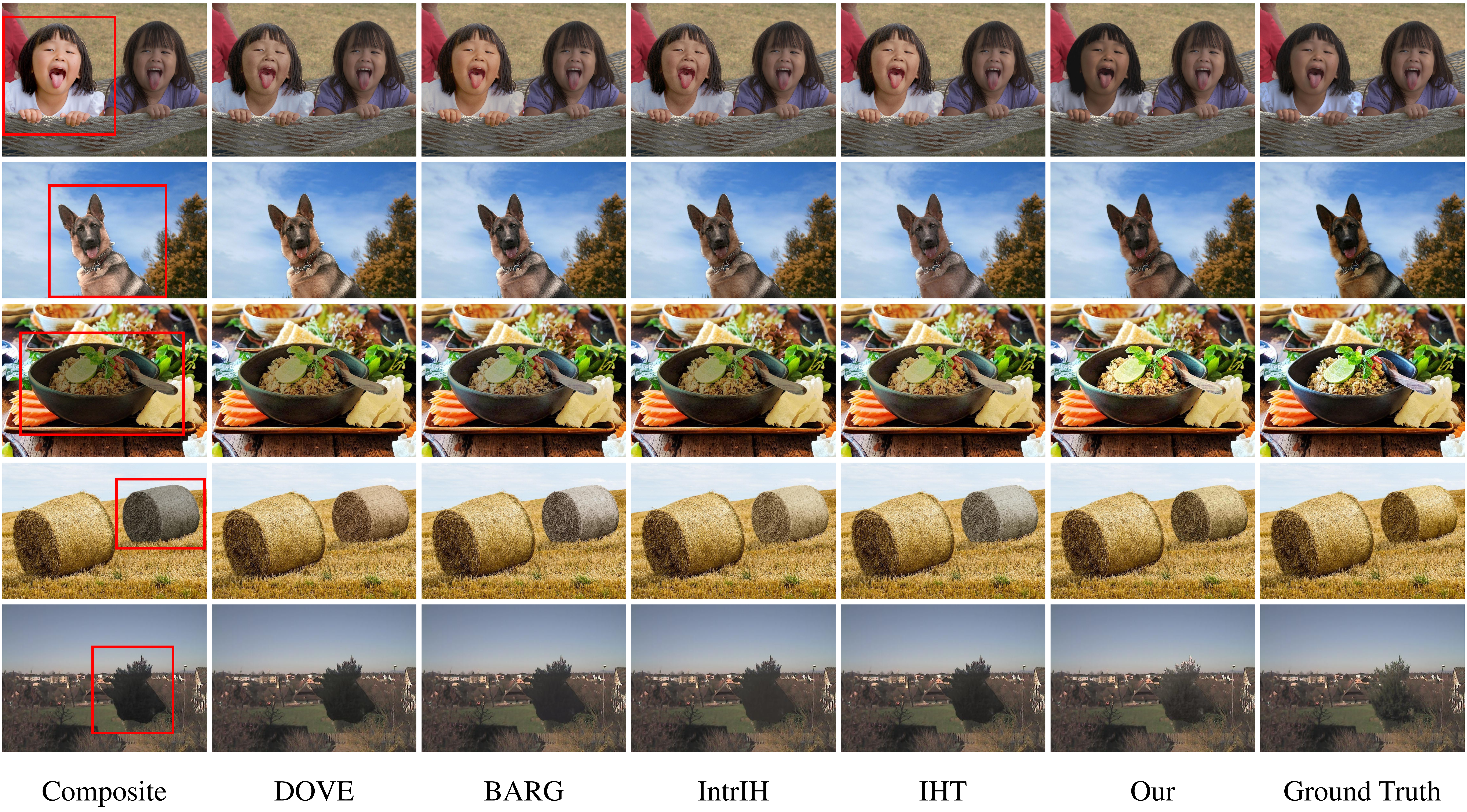}
{\begin{center}
\vspace{-0.5cm}
\caption{\textbf{Visual Comparison on iHarmony4.} The red boxes in the composite images indicate the foreground regions. Zoom in for better visualization.}
\label{fig:visual}
\end{center}
}
\vspace{-0.7cm}
\end{figure}

For practical applications, which typically use higher image resolutions, the quantitative results at $256\times256$ resolution as shown above may not reflect the actual image harmonization performance. To study this issue, we further evaluate Harmonizer with the strong baseline DOVE and the state-of-the-art BARG/IHT at high resolutions. We compute the metrics at the original resolutions of the images in iHarmony4. Note that the subsets in iHarmony4 have different resolutions, {\it e.g.}, the average image size for HCOCO is about $500\times500$ and for HAdobe5k is about $3000\times3000$.
Since high-resolution inputs would significantly degrade the performances of the existing methods \ke{that are trained on $512\times512$ resolution}, we still input low-resolution images to them and then bilinearly upsample their results to high resolutions for metric calculation. \ke{To avoid blurry outputs caused by bilinear upsampling,}
we also upsample their results using Polynomial Color Mapping (PCM)~\cite{afifi2020deepWB}, which can transfer the foreground appearances in the low-resolution outputs to the high-resolution composite images without loss of details (has been used for visualization in previous works). In contrast, our Harmonizer can process the high-resolution inputs directly as its filters are resolution-independent.
Table\,\ref{tab:highresolution} shows the metrics computed at the original image resolutions. The performances of the existing methods are significantly lowered, {\it e.g.}, MSE/fMSE of IHT are increased from 37.07/395.66 to 57.22/741.98. In contrast, Harmonizer only has a small performance drop at high resolutions, and its MSE is now $50\%$ lower than the state-of-the-art method. We provide some visual comparisons in Fig.\,\ref{fig:visual}.

Table\,\ref{tab:other} compares different methods in terms of inference speed, model size, and memory requirement. A fast inference speed is necessary for real-time applications, while a small model size and a low memory requirement facilitate deployment on mobile devices. 
Our results demonstrate that Harmonizer is faster, lighter, and more memory efficient than other methods. Remarkably, on a RTX3090 GPU, Harmonizer can process a 1080P (Full HD) video at 56 {\it fps}, about $4\times$ faster than the recent fastest method DOVE~\cite{DoveNet}. Moreover, we observe that Harmonizer can be further accelerated by a fusion implementation of filters or using techniques like Halide Auto-Scheduler.

\subsection{Video Harmonization Results}\label{sec:4_2}

\begin{figure}[t]
\centering
\includegraphics[width=0.99\linewidth]{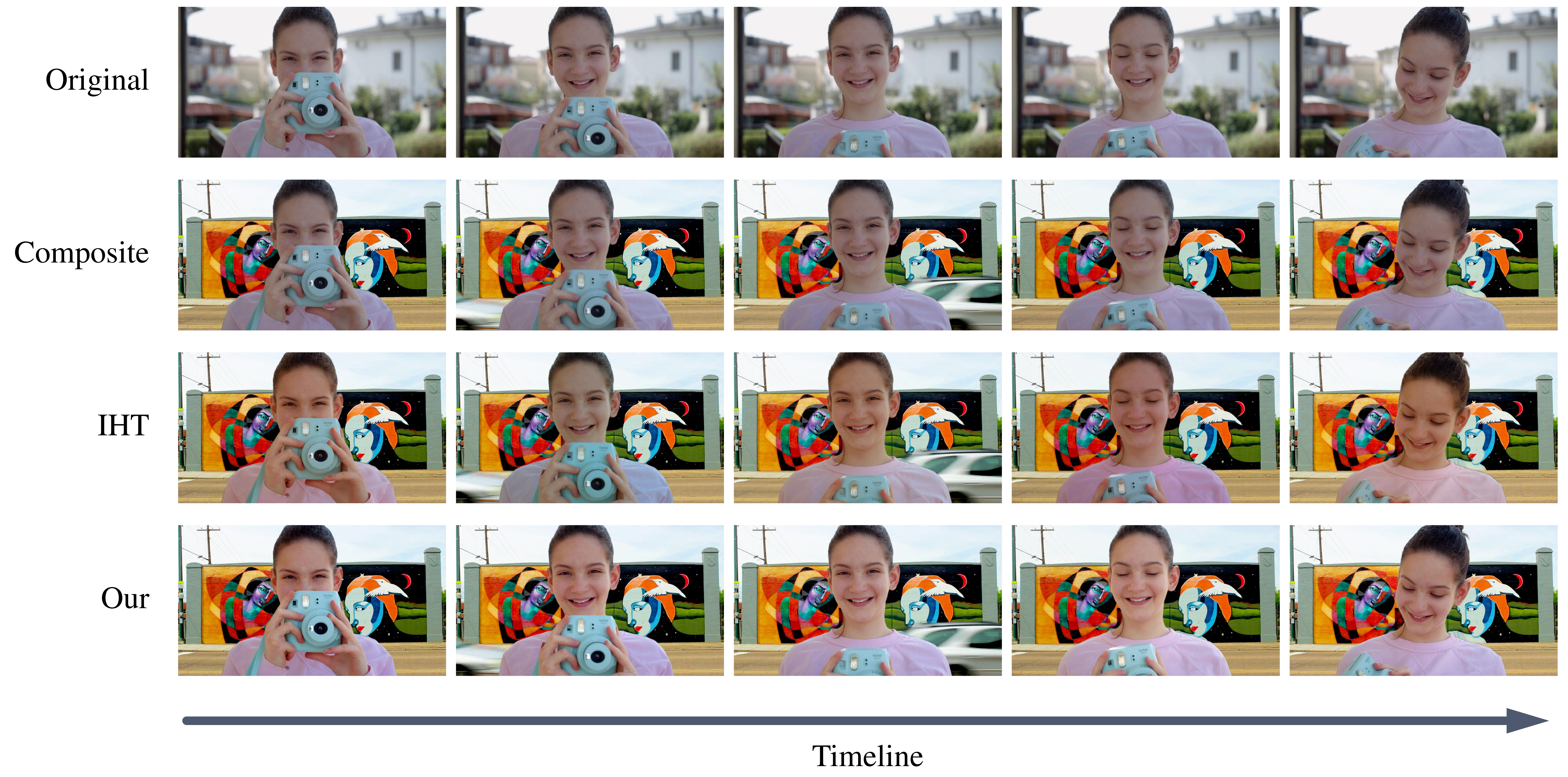}
{\begin{center}
\vspace{-0.5cm}
\caption{\textbf{Video Harmonization Results.} Previous image harmonization methods (we only visualize the results from IHT\,\cite{TransformerIH} here due to space limitation) output the frames with inconsistent foreground appearances ({\it e.g.}, the cloth regions). Instead, our Harmonizer obtains consistent harmonization results across frames.
}
\label{fig:video}
\end{center}
}
\vspace{-0.5cm}
\end{figure}

As shown in Fig.\,\ref{fig:video}, by applying Eq.\,\ref{eq:video}, 
Harmonizer obtains stable harmonization results across video frames.
On the contrary, the results of the prior methods suffer from severe flickers. Unfortunately, our strategy proposed in Sec.\,\ref{sec:video} is not suitable for use in prior existing methods since they solve harmonization as a pixel-wise image translation problem in a black-box manner.

\begin{table*}[t]\label{tab:ablation}
  \begin{center}
      \caption{\textbf{Ablation of Harmonizer on iHarmony4.} {\it Left}: Evaluating the effectiveness of different Harmonizer components at 256x256 resolution. {\it Right}: Evaluating the performance of Harmonizer with different numbers of filters. MSE and PSNR are calculated at 256x256 resolution, while {\it fps} is measured at 1080P resolution.}\label{tab:ablation}
      \vspace{-0.3cm}
  \begin{minipage}[t]{.53\textwidth}
    \vspace{0pt}
    \centering
    \setlength{\tabcolsep}{4pt}
    \scriptsize
    \begin{tabular}{cc|P{2.3}P{2.3}}
      \toprule 
      Cascade & Dynamic & \multicolumn{2}{c}{Metrics} \\ 
    Regressor & Loss Strategy & \multicolumn{1}{c}{$\text{MSE}\downarrow$} & \multicolumn{1}{c}{$\text{PSNR} \uparrow$} \\
      \midrule
      & & 28.47 &  36.96 \\
      \checkmark & & 26.85 &  37.23 \\
      \checkmark & \checkmark & \textbf{24.26} & \textbf{37.84}  \\
      \bottomrule
    \end{tabular}
  \end{minipage}%
  \begin{minipage}[t]{.47\textwidth}
    \vspace{0pt}
    \centering
        \setlength{\tabcolsep}{4pt}
    \scriptsize
    \begin{tabular}{r|rrrr}
      \toprule 
      Metrics & \multicolumn{4}{c}{Number of Filters} \\
      & \multicolumn{1}{c}{2} & \multicolumn{1}{c}{4} & \multicolumn{1}{c}{6} & \multicolumn{1}{c}{8} \\
      \midrule
      $\text{MSE}\downarrow$ & 74.16 & 29.60 & 24.26 & \textbf{23.51} \\
      $\text{PSNR}\uparrow$ & 33.49 & 36.75 & 37.84 & \textbf{38.06} \\
       $\textit{fps}\uparrow$ & \textbf{86.2} & 63.5 & 56.3 & 51.9 \\
      \bottomrule
    \end{tabular}
  \end{minipage}
  \vspace{-0.5cm}
  \end{center}
\end{table*}

\subsection{Ablation Study}\label{sec:4_3}
In Table\,\ref{tab:ablation}\,\textit{Left}, we evaluate the cascade regressor and dynamic loss strategy proposed in Harmonizer. The results show that both techniques can improve the image harmonization performance. We also observe that even without these two techniques, the results of Harmonizer (MSE of $33.28$) still surpass the previous state-of-the-art (MSE of $37.07$), which demonstrates that the overall design of Harmonizer has advantages over the black-box autoencoders. 
In Table\,\ref{tab:ablation}\,\textit{Right}, we analyze the impact of different numbers of filters on the performance and speed. 
Specifically, we validate Harmonizer with 2, 4, 6, and 8 filters. 
For the experiments with 6 or 8 filters, we choose the most important filters based on Fig.\,\ref{fig:userstudy}. For the experiments with 2 or 4 filters, we select the filters randomly and report the metrics averaged over 3 runs. The results show that the performance of Harmonizer increases as the number of filters increases. Besides, Harmonizer needs at least 4 filters to avoid large performance degradation.

 \begin{figure}[t]
\centering
\includegraphics[width=0.99\linewidth]{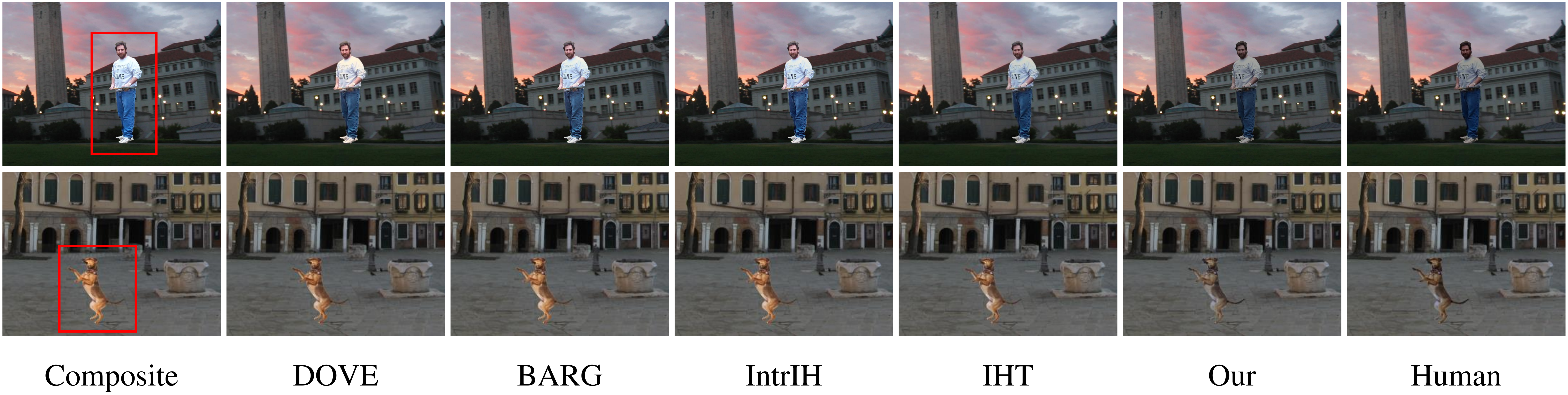}
{\begin{center}
\vspace{-0.5cm}
\caption{\textbf{Visual Comparison on Real Composite Images.} ``Human'' means that the results are produced by humans.}
\label{fig:realworld}
\end{center}
}
\vspace{-0.4cm}
\end{figure}

\subsection{User Studies}\label{sec:4_4}

We show the advantage of Harmonizer in real-world image/video harmonization through user studies. For image harmonization, we use the real composite images released in \cite{DIH}, which includes 99 images. Since these images have no labels, we ask a skilled human to process them for reference. For video harmonization, we composite the 20 foreground videos generated by the video matting methods~\cite{Qin_2020_PR,MODNet,lin2021robust} with 20 new background videos to create \ke{composite} videos (one of them is shown in Fig.\,\ref{fig:video}). 
We invite 12 participants to rank the results from different methods and the human. In Table\,\ref{tab:b-t}, we follow prior works to use the Bradley-Terry model (B-T model)~\cite{btmodel} for ranking.
Harmonizer achieves the highest B-T scores. The B-T scores of applying prior methods to video harmonization are even lower than the original composite inputs due to severe flickering.
Fig.\,\ref{fig:realworld} visualizes two samples used in our image harmonization user study.

\begin{table*}[t]
  \begin{center}
    \caption{\textbf{User Study Results.} We calculate B-T scores to quantify our user study results. For image harmonization, the results from humans are compared.}\label{tab:b-t}
\setlength{\tabcolsep}{4pt}
\scriptsize
\begin{tabular}{r|rrrrr|rr}
      \toprule 
      Metrics & Composite & DOVE\cite{DoveNet} & BARG\cite{BargainNet} & IntrIH\cite{IntrinsicIH} & IHT\cite{TransformerIH} & Our & Human \\
      \midrule
      B-T Score (Image) $\uparrow$ & 0.412 & 0.639 & 0.618 & 0.663 & 0.724 & \uline{1.028} & \textbf{1.393} \\
      B-T Score (Video) $\uparrow$ & \uline{1.173} & 0.587 & 0.497 & 0.530 & 0.601 & \textbf{2.042} & - \\
      \bottomrule
    \end{tabular}
\vspace{-0.5cm}
\end{center}
\end{table*}

 \begin{figure}[ht]
\centering
\vspace{-0.5cm}
\includegraphics[width=0.99\linewidth]{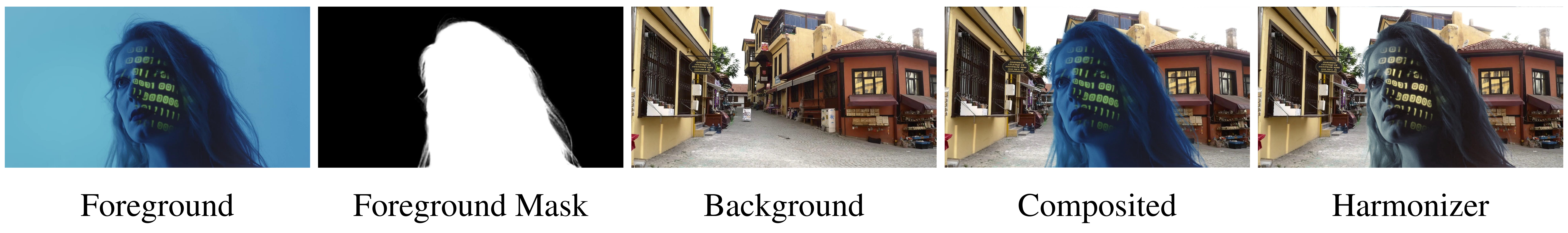}
{\begin{center}
\vspace{-0.5cm}
\caption{\textbf{A Failure Case of Harmonizer.} We show a composite image with a strong hue difference (within the blue color range) between foreground and background.}
\label{fig:fail_visual}
\end{center}
}
\vspace{-1.5cm}
\end{figure}

\section{Conclusion}

In this paper, we have studied the image harmonization process carried out by humans, which has inspired us to design Harmonizer. Benefited by our novel architecture and two proposed techniques, {\it i.e.}, a cascade regressor and a dynamic loss strategy, Harmonizer is lighter and faster than prior methods while achieving new state-of-the-art performances. Nevertheless, our method does have limitations. \ke{It may fail to handle color-specific appearance inconsistency or the different lighting conditions between the foreground and background. Fig.\,\ref{fig:fail_visual} shows one case.} As a future work, we would like to develop more complex image filters, {\it e.g.}, color-separated filters, to address the problem.

\clearpage

 \begin{figure}[t]
\centering
\includegraphics[width=0.99\linewidth]{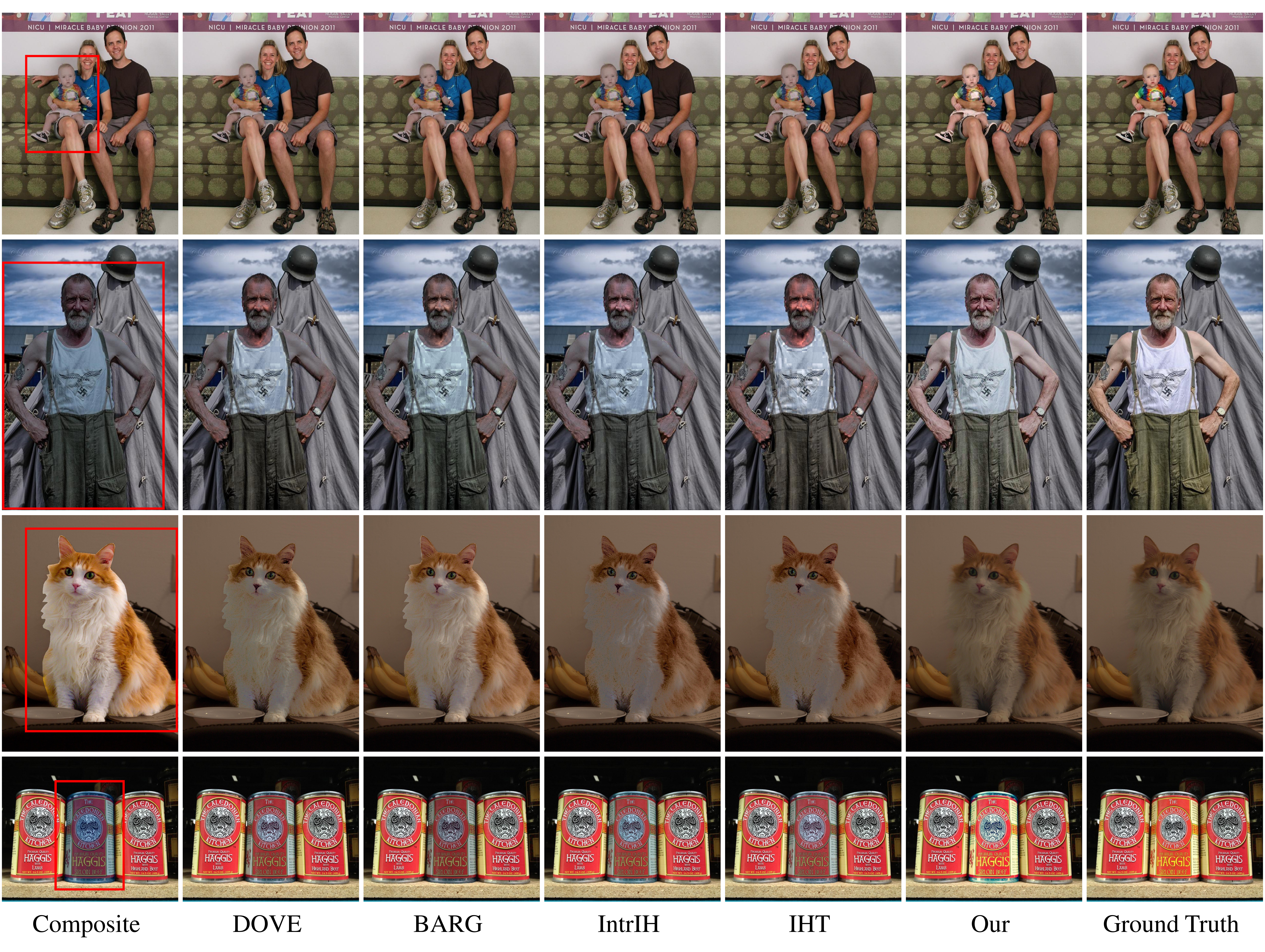}
{\begin{center}
\vspace{-0.5cm}
\caption{\textbf{Visual Comparison on iHarmony4.} The red boxes in the composite images indicate the foreground regions. Zoom in for better visualization.}
\label{fig:more_visual}
\end{center}
\vspace{-0.5cm}
}
\end{figure}

 \begin{figure}[h]
\centering
\includegraphics[width=0.99\linewidth]{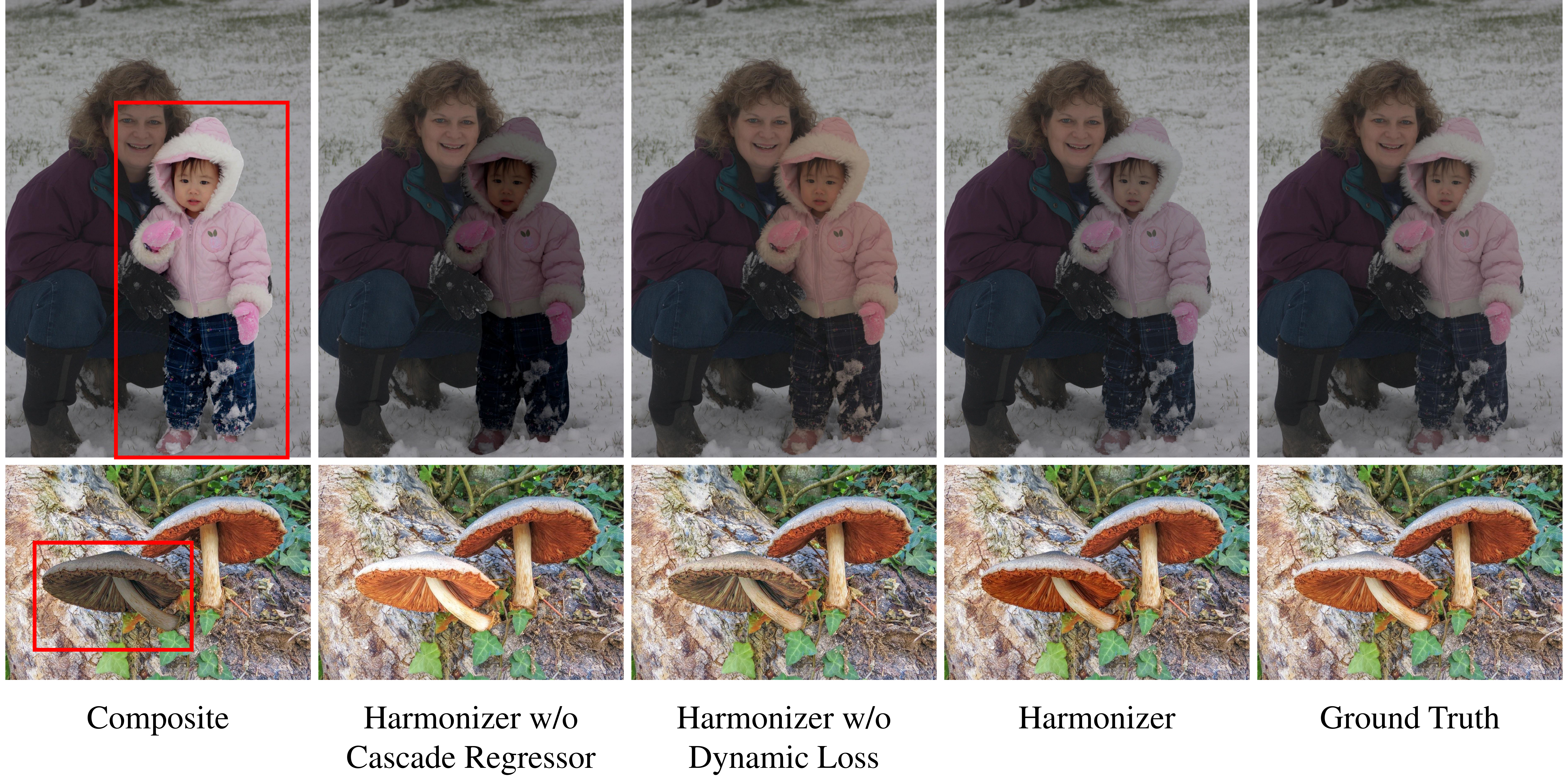}
{\begin{center}
\vspace{-0.5cm}
\caption{\textbf{Visual Results of Ablation Experiments.} We show the Harmonizer results without cascade regressor or dynamic loss strategy. Without the cascade regressor, the filters in Harmonizer will affect each other, {\it e.g.}, the too bright or too dark results may be caused by adjusting brightness/highlight/shadow simultaneously (the 2\textit{nd} column). Without the dynamic loss strategy, Harmonizer will bias toward some filters, {\it e.g.}, the predicted color temperature argument may be inaccurate (the 3\textit{rd} column).}
\label{fig:ablation_visual}
\end{center}
\vspace{-0.5cm}
}
\end{figure}

\clearpage
%
%
\bibliographystyle{splncs04}
\bibliography{egbib}
\end{document}